%% file: main.tex
\documentclass[journal]{IEEEtran}

\usepackage[utf8]{inputenc} 
\usepackage[T1]{fontenc}
\usepackage{amsmath,amssymb,amsfonts}
\usepackage{algorithmic}
\usepackage{graphicx}
\usepackage{textcomp}
\usepackage{xcolor}
\usepackage{tikz} 
\usetikzlibrary{shapes} 
\usetikzlibrary{calc}
\usetikzlibrary{positioning} 
\usepackage{caption}
\usepackage{subcaption}
\usepackage{booktabs} 
\usepackage{multirow}
\usepackage{url}
\usepackage{cite}
\usepackage{longtable}
\usepackage{hyperref}
\usepackage{listings} 

\begin{document}

\title{Query as Test: An Intelligent Driving Test and Data Storage Method for Integrated Cockpit-Vehicle-Road Scenarios}

\author{
        Shengyue Yao$^{\dag}$,
        RunQing Guo$^{\dag}$,
        Yangyang Qin,
        Xiangbin Meng,
        Jipeng Cao,
        Yilun Lin$^{\star}$, \\
        Yisheng Lv,
        Li Li,
        and Fei-Yue Wang


\thanks{$\dag$ Equal contribution.}
\thanks{$\star$ Corresponding author: Yilun Lin(lin.yilun@h2oslabs.com).}

\thanks{Shengyue Yao, Jipeng Cao, Yilun Lin,  (\{yao.shengyue, cao.jipeng, lin.yilun\}@h2oslabs.com) is with Peking University International Innovation Center, Lin-gang Special Area (PKU-IICSH), Shanghai 201306, China, and also with Onesyn (Shanghai) Technology Co., Ltd, Shanghai 201306, China.}%

\thanks{RunQing Guo, Yangyang Qin (\{guorunqing, qinyangyang\}@catarc.ac.cn) is with CATARC Automotive Technology(Shanghai) Co.,Ltd, Shanghai 201800, China.}%

\thanks{Xiangbin Meng (Xiangbing.Meng1@geely.com) is with ZEEKR Intelligent Technology Holding Limited, Ningbo 315000, Zhejiang Province, China.}

\thanks{$^{4}$Li Li (li-li@tsinghua.edu.cn) is with the Department of Automation, Tsinghua University, Beijing 100084, China.}

\thanks{Fei-Yue Wang, Yisheng Lv (\{feiyue.wang, lv.yisheng\}@ia.ac.cn) is with the State Key Laboratory for Management and Control of Complex Systems, Chinese Academy of Sciences, Beijing 100190, China. Fei-Yue Wang is also with the Macao Institute of Systems Engineering, Macau University of Science and Technology, Macao 999078, China.}









}

\markboth{IEEE Transactions on Intelligent Transportation Systems, 2025}%
{Yao \MakeLowercase{\textit{et al.}}: Query as Test for Intelligent Driving}

\maketitle

\begin{abstract}
With the deep penetration of Artificial Intelligence (AI) in the transportation sector, intelligent cockpits, autonomous driving, and intelligent road networks are developing at an unprecedented pace. However, the data ecosystems of these three key areas are increasingly fragmented and incompatible, severely hindering system collaboration, advanced semantic querying, safety validation, and privacy protection. To address the issues that existing testing methods rely on data stacking, fail to cover all edge cases, and lack flexibility, this paper first introduces the concept of "Query as Test" (QaT). This paradigm shifts the focus from rigid, prescripted test cases to flexible, on-demand logical queries against a unified data representation. To support this new paradigm, we identify the need for a fundamental improvement in data storage and representation, leading to our proposal of "Extensible Scenarios Notations" (ESN). ESN is a novel declarative data framework based on Answer Set Programming (ASP), which uniformly represents heterogeneous multimodal data from the cockpit, vehicle, and road as a collection of logical facts and rules. This approach not only achieves deep semantic fusion of data, but also brings three core advantages: First, it supports complex and flexible semantic querying through logical reasoning; Second, its inherent logical structure provides natural interpretability for decision-making processes; Third, it allows for on-demand data abstraction through logical rules, enabling fine-grained privacy protection. We further elaborate on the QaT paradigm, transforming the functional validation and safety compliance checks of autonomous driving systems into logical queries against the ESN database, significantly enhancing the expressiveness and formal rigor of the testing. Finally, we introduce the concept of "Validation-Driven Development" (VDD), proposing that, in the era of Large Language Models (LLMs), development should be guided by logical validation rather than being limited by traditional quantitative testing, thereby accelerating the iteration and development process.
\end{abstract}


\input{contents/01-introduction}
\input{contents/02-related-works}
\input{contents/03-esn}
\input{contents/04-qat}
\input{contents/05-experiments}
\input{contents/06-discussion}
\input{contents/07-conclusion}


\input{main.bbl}
\clearpage
\newpage
\section*{Acknowledgment}
The authors would like to express their sincere gratitude to all contributors who made this work possible. The initial concept of "Query as Test" (QaT) and the Extensible Scenarios Notations (ESN) framework was conceived through fruitful discussions among Yilun Lin, Shengyue Yao, Xiangbin Meng, RunQing Guo, and Jipeng Cao. The manuscript was primarily authored by Yilun Lin, Shengyue Yao, and Xiangbin Meng, with the implementation of the ESN framework carried out by Shengyue Yao and Yilun Lin. RunQing Guo, Yangyang Qin, and Jipeng Cao provided invaluable scenario data support and industrial application requirements that significantly enhanced the practical relevance of the QaT paradigm. Yisheng Lv, Li Li, and Fei-Yue Wang offered insightful discussions and provided critical suggestions for manuscript revisions that greatly improved the quality and clarity of this work. The authors also acknowledge the support from their respective institutions and the collaborative research environment that facilitated this interdisciplinary study on intelligent driving validation and data representation.

\clearpage
\newpage
\appendices
\input{appendices/i-scenarios}
\clearpage
\newpage
\input{appendices/ii-query.tex}

\end{document}

%% file: contents/01-introduction.tex
\section{Introduction}
\label{sec:introduction}

The field of transportation is undergoing a profound transformation driven by Artificial Intelligence (AI), advancing along three parallel yet interconnected axes: \textbf{Intelligent Cockpits (IC)}, \textbf{Autonomous Driving (AD)}, and \textbf{Vehicle-to-Everything (V2X)}~\cite{Chen-2023, Wang-2017a}. However, despite significant progress in each domain, their development paths have largely remained isolated, lacking a unified data and semantic "glue". This fragmentation presents a critical challenge: the data formats, modalities, and semantics of cockpit interactions, vehicle dynamics, and roadside infrastructure are vastly different~\cite{Ounoughi-2023}. This separation not only limits the synergistic potential among the three domains but also creates bottlenecks for the safety, reliability, and efficiency of the entire transportation system.

A primary manifestation of this problem is found in the testing and validation of autonomous driving systems~\cite{Li-2018, Li-2019}. Current testing methodologies predominantly rely on a paradigm of data stacking, where massive volumes of driving data are collected and replayed in simulations or field tests to identify system failures. This approach, while valuable, suffers from fundamental limitations. Firstly, it is inherently reactive and inefficient, requiring enormous datasets to cover even a fraction of possible driving scenarios. Secondly, and more critically, it struggles to flexibly and exhaustively cover all edge cases, rare but critical events that often lead to accidents~\cite{Li-2022a}. The process of creating test cases is rigid, making it difficult to dynamically query for complex, behavior-driven scenarios. For instance, one cannot easily query a petabyte-scale dataset for "all instances where a vehicle performed an emergency brake maneuver in rainy weather due to a pedestrian jaywalking while the driver was distracted."

To circumvent these limitations, this paper introduces a new paradigm: \textbf{"Query as Test" (QaT)}. The core idea of QaT is to shift the testing philosophy from executing a fixed set of test cases to performing dynamic, expressive logical queries against a unified representation of scene data. The paradigm's innovation is twofold. Firstly, instead of asking "Did the system pass test case X?", we ask "Can we find any instance in the data that violates safety property Y?". This transforms testing from a brute-force data coverage problem into a sophisticated discovery process for unforeseen failure modes. Secondly, and more profoundly, QaT enables asking \textbf{what-if} questions. By altering the logical rules of the query itself, one can perform on-the-fly counterfactual analysis without needing to re-run simulations. For instance, one could query "Would a collision have been avoided if the braking policy were more aggressive?" by temporarily modifying the rules that define the vehicle's braking behavior. This elevates testing from mere discovery to active exploration, offering unparalleled flexibility and deeper causal insights into system behavior.

However, the realization of the QaT paradigm is contingent on a fundamental revolution in how we store and represent traffic scene data. Existing data representations, which range from extremely specific numerical datasets (e.g., Waymo Open Dataset~\cite{Sun-2020}, nuScenes~\cite{Caesar-2020}) to overly abstract semantic descriptions, are ill-suited for the complex logical querying that QaT demands. To address this foundational gap, we propose the \textbf{Extensible Scenarios Notation (ESN)}.

ESN is a declarative framework that leverages \textbf{Answer Set Programming (ASP)}---a powerful form of declarative logic programming---to serve as a canonical representation for all scene data. By converting heterogeneous data from cockpits, vehicles, and roads into a collection of ASP facts and rules, we create a unified, logical knowledge base. This ESN framework offers four key advantages:
\begin{enumerate}
    \item \textbf{Semantic Querying}: Complex, event-based queries can be answered through logical inference.
    \item \textbf{Inherent Interpretability}: The reasoning process for any query result provides a clear, human-readable explanation.
    \item \textbf{Flexible Privacy Control}: Information granularity can be adjusted dynamically through logical abstraction rules.
    \item \textbf{Exceptional Extensibility}: New data sources, concepts, or events can be added by simply introducing new facts and rules.
\end{enumerate}

Based on this foundation, the "Query as Test" paradigm allows safety specifications and functional requirements to be formalized as logical queries over the ESN database, significantly enhancing the rigor and expressiveness of the validation process.

This paper is structured as follows: Section \ref{sec:related-works} reviews related work. Section \ref{sec:esn} details the ESN methodology. Section \ref{sec:qat} elaborates on the "Query as Test" paradigm. Section \ref{sec:experiments} presents the experimental design. Section \ref{sec:discussion} discusses the implications, including the concept of "Validation-driven Development", and future directions. Section \ref{sec:conclusion} concludes the paper.

%% file: contents/02-related-works.tex
\section{Related Work}
\label{sec:related-works}

This section examines existing research across three critical domains: autonomous driving data standards, logical knowledge representation, and system validation. Our analysis reveals a convergent trend toward bridging the gap between low-level physical data and high-level semantic understanding---a challenge that motivates our ESN-QaT framework.

\subsection{Autonomous Driving Data and Testing Paradigms}

Current autonomous driving datasets, while successful for machine learning applications, exhibit fundamental limitations for semantic analysis. Large-scale datasets such as Waymo Open Dataset~\cite{Sun-2020} and nuScenes~\cite{Caesar-2020} are optimized for numerical processing rather than logical querying, making it challenging to identify complex behavioral patterns. For instance, detecting "aggressive lane-changing" scenarios requires extensive post-processing of trajectory data rather than direct semantic queries.

In parallel, scenario description standards like ASAM OpenSCENARIO~\cite{ASAM-OpenSCENARIO-2020} focus primarily on test case \textit{generation} rather than data \textit{analysis}. This creates a critical gap in validation workflows: while OpenSCENARIO can generate parameterized test variants, there lacks a systematic approach to discover meaningful scenarios from real-world data. Our ESN framework addresses this complementary need, enabling the discovery of high-risk scenarios that can subsequently inform OpenSCENARIO-based test generation.

Furthermore, current testing approaches face the fundamental \textbf{test oracle problem}---the difficulty of defining meaningful pass/fail criteria. Traditional numerical assertions (e.g., Time-to-Collision thresholds) often fail to capture nuanced aspects of safe and comfortable driving behavior, motivating our shift toward principle-based logical validation.

\subsection{Logical Knowledge Representation for Dynamic Systems}

The representation of temporal and dynamic knowledge has been extensively studied in artificial intelligence. Classical approaches like Situation Calculus~\cite{McCarthy-1963} and Event Calculus~\cite{Shanahan-1995} provide theoretical elegance but lack the computational efficiency required for large-scale data processing. While Datalog~\cite{Green-2013} offers more practical recursive querying capabilities, it cannot handle the non-monotonic reasoning essential for modeling traffic scenarios with exceptions and defaults.

Answer Set Programming (ASP)~\cite{Gelfond-1988} emerges as a compelling solution to these limitations. Through its stable model semantics, ASP naturally handles default reasoning, exceptions, and incomplete information---characteristics fundamental to real-world driving scenarios. The proven applicability of ASP in complex domains such as planning and robotics provides strong theoretical foundation for its adoption as ESN's logical backbone.

\subsection{Validation, Interpretability, and Trust}

As autonomous systems become increasingly complex, ensuring their interpretability and trustworthiness becomes paramount~\cite{Olaverri-Monreal-2020a}. Current explainable AI approaches predominantly rely on post-hoc methods~\cite{Ribeiro-2016, Lundberg-2017} that approximate model behavior, potentially leading to misleading explanations. In contrast, ESN provides inherent interpretability through formal proof traces, offering verifiable explanations crucial for safety-critical applications.

Similarly, privacy-preserving data sharing presents ongoing challenges. While differential privacy~\cite{Dwork-2006} provides strong theoretical guarantees, its utility-privacy trade-offs often limit practical applicability. ESN's logical abstraction offers an alternative paradigm, enabling semantic-level privacy control through the sharing of high-level conclusions rather than raw data.

Finally, traditional formal verification methods~\cite{Clarke-1997, Clarke-1981}, though providing the highest assurance levels, face scalability limitations when applied to complete autonomous driving systems. Our Query as Test paradigm addresses this challenge by applying formal reasoning principles to concrete execution data rather than abstract models, making rigorous validation more tractable for complex AI systems.

%% file: contents/03-esn.tex
\section{Extensible Scenarios Notation: From Raw Data to Logical Reasoning}
\label{sec:esn}

This section details the core principles of the Extensible Scenarios Notation (ESN). The design goal of the ESN is to transform and unify heterogeneous, multimodal data from the cockpit, vehicle, and road domains into a formal, logical representation based on Answer Set Programming (ASP)~\cite{Gelfond-1988}. This process is termed "ASP-ification." We will describe ESN's core formalism, the preprocessing pipeline for mapping datasets like WOMD~\cite{Sun-2020} into ESN facts, the library for defining high-level semantic events, and the interoperability layer for cross-domain data fusion.

\begin{figure}[htbp]
    \centering
    \includegraphics[width=0.9\columnwidth]{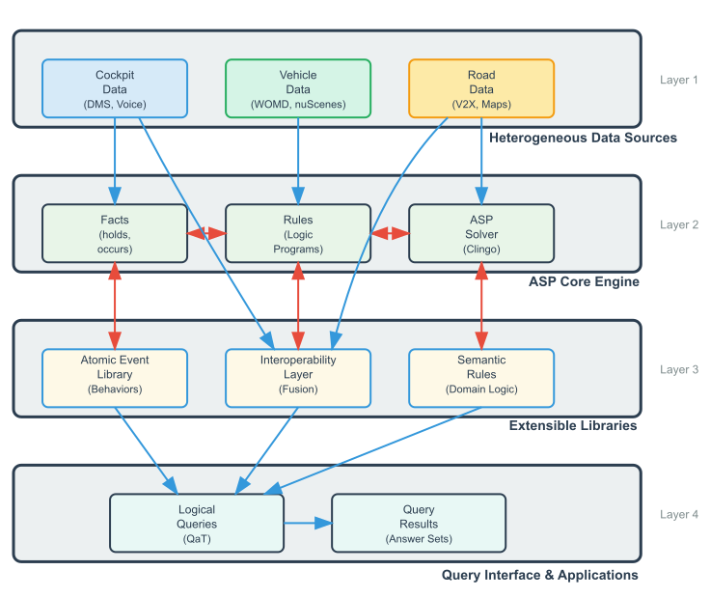}
    \caption{The ESN framework architecture organized in four layers: (1) Heterogeneous data sources from cockpit, vehicle, and road domains; (2) ASP core engine with facts, rules, and solver; (3) Extensible libraries for semantic reasoning and cross-domain fusion; (4) Query interface supporting the QaT paradigm.}
    \label{fig:esn_framework}
\end{figure}

\subsection{ESN Core Formalism: Representing Scenarios as Answer Set Programs}

In ESN, a traffic scenario is defined as an Answer Set Program---a collection of logical facts and rules. The core of this representation lies in its spatio-temporal logic predicates, which are designed based on established practices in robotics and dynamic systems reasoning~\cite{McCarthy-1963, Shanahan-1995}.

We define two fundamental predicates to describe the dynamic evolution of a scene:
\begin{itemize}
    \item \texttt{holds(Fluent, Timestamp)}: This predicate states that a "fluent" (a property that persists over time) is true at a discrete \texttt{Timestamp}. Fluents represent the continuous state of the system at a given moment.
    
    \textit{Example:} \texttt{holds(position(car\_01, 15.2, 45.8, 0.5), 1622541987.1).} This example means, at the moment corresponding to timestamp $1622541987.1$, the vehicle with ID $car\_01$ has a position state of coordinates $(15.2, 45.8, 0.5)$.
    
    \item \texttt{occurs(Event, Timestamp)}: This predicate states that an "event" (an instantaneous action or observation) happens at a \texttt{Timestamp}. Events represent actions that can cause a change in state.
    
    \textit{Example:} \texttt{occurs(brake\_light\_on(car\_02), 1622541987.2).} This example means, At the moment corresponding to timestamp $1622541987.2$, the event 'brake light on' occurred for the vehicle with ID $car\_02$.
\end{itemize}
A basic example of scene representation in ESN is demonstrated in Figure \ref{list:esn_example}.

\begin{figure}[!ht]
    \centering
    \begin{lstlisting}
% ESN Example: Basic Scene Representation
% Vehicle facts
holds(position(car_01, 15.2, 45.8, 0.5), 1622541987.1).
holds(velocity(car_01, 25.5, 0.0, 0.0), 1622541987.1).
holds(position(car_02, 18.7, 45.8, 0.5), 1622541987.1).
holds(velocity(car_02, 30.2, 0.0, 0.0), 1622541987.1).

% Traffic light state
holds(traffic_light_state(lane_001, red), 1622541987.1).

% Driver state
holds(driver_state(car_01, attentive), 1622541987.1).

% Event
occurs(brake_light_on(car_02), 1622541987.2).

% Rule: Following relationship
is_following(V1, V2, T) :-
    holds(position(V1, X1, Y1, Z1), T),
    holds(position(V2, X2, Y1, Z1), T),
    X1 < X2,
    distance(V1, V2, T, D),
    D < 50.0.

% Rule: Distance calculation
distance(V1, V2, T, D) :-
    holds(position(V1, X1, Y1, Z1), T),
    holds(position(V2, X2, Y2, Z2), T),
    D = sqrt((X2-X1)^2 + (Y2-Y1)^2 + (Z2-Z1)^2).
    \end{lstlisting}
    \caption{Example ESN representation showing facts, events, and rules for a traffic scenario. Note the example code has been simplified for better understanding.}
    \label{list:esn_example}
\end{figure}

Additionally, ESN includes static predicates describing the background context of the scene, such as map topology or vehicle properties. A complete ESN scene consists of a vast number of such facts. Querying an ESN scene is equivalent to posing a logical question to this fact and rule base. The ASP solver's task is to find one or more "answer sets," each being a logically consistent set of facts that represents a valid solution to the query.

\subsection{Preprocessing Pipeline: The ASP-ification of Heterogeneous Data}
Converting raw data into the ESN format is a critical first step. Our goal is to propose a general methodology for this transformation that is applicable across different data sources. We have designed a set of principles to guide this "ASP-ification" process for the key domains of an integrated driving scenario.

\subsubsection{General Principles of Data Transformation}
The transformation process follows a set of rules to map real-world data into the ESN's logical predicates (\texttt{holds}, \texttt{occurs}, and static facts).

\begin{itemize}
    \item \textbf{Vehicle Data}: This is the core data stream, describing the physical state and properties of traffic participants. Dynamic data, such as trajectories, velocity, and acceleration at each timestamp, are mapped to \texttt{holds} predicates (e.g., \texttt{holds(position(V, X, Y), T)}). Static or semi-static properties, like vehicle dimensions or object class (car, pedestrian), are represented as timeless facts (e.g., \texttt{object\_property(V, dimension, L, W, H)}).

    \item \textbf{Cockpit Data}: This domain captures interactions within the vehicle. Continuous states from the Driver Monitoring System (DMS), such as the driver being "distracted" or "attentive," are mapped to \texttt{holds} predicates (e.g., \texttt{holds(driver\_state(attentive), T)}). Instantaneous events, like a passenger giving a voice command or touching the infotainment screen, are mapped to \texttt{occurs} predicates (e.g., \texttt{occurs(voice\_command("navigate home"), T)}).

    \item \textbf{Roadside Data}: This includes data from intelligent infrastructure. Static map elements like lane boundaries, stop lines, and traffic signs are converted into a set of static facts that define the scene's topology (e.g., \texttt{map\_lane(lane\_12, ...)}). Dynamic information from V2X messages, such as Signal Phase and Timing (SPaT) data, is converted into \texttt{holds} predicates (e.g., \texttt{holds(traffic\_light\_state(tl\_A, red), T)}), while warning messages (e.g., for an accident ahead) are mapped to \texttt{occurs} predicates.
\end{itemize}

\subsubsection{Adaptation for Common Public Datasets}
The general transformation principles provide a universal blueprint that can be readily instantiated for prominent public datasets. Although these benchmarks primarily focus on vehicle and road data, they offer a rigorous testing ground for a significant portion of our framework.

\begin{itemize}
    \item \textbf{Waymo Open Motion Dataset (WOMD)}~\cite{Sun-2020}: The ESN transformation is instantiated by creating a specific parser for WOMD's official \texttt{Scenario} Protocol Buffer definition. This parser systematically processes each \texttt{Scenario} within a \texttt{.tfrecord} file, translating its constituent parts—object tracks, map features, and dynamic states—into a cohesive set of ESN facts, thereby creating a self-contained, logical scene representation.

    \item \textbf{nuScenes Dataset}~\cite{Caesar-2020}: For nuScenes, with its relational database schema, the instantiation involves mapping its tables directly to ESN facts. Records from tables such as \texttt{instance} and \texttt{sample\_annotation} are logically transformed into object properties and timestamped \texttt{holds} predicates for attributes like position and dimension.
\end{itemize}

This adaptability demonstrates that the ESN pipeline is not merely conceptual but is a concrete, structured process capable of unifying diverse data sources, from real-world sensors to standard industry benchmarks, into a single, queryable logical framework.

\subsection{Atomic Event Library: Defining Dynamic Behaviors with ASP Rules}

The power of ESN lies in its ability to derive high-level semantic events from these base facts. This is achieved through an extensible library of ASP rules that define "atomic events." These events are the building blocks for more complex scenario logic.

For example, an \texttt{is\_turning} event can be defined by a rule that checks if a vehicle's heading changes by more than a certain threshold between consecutive timestamps. Similarly, an \texttt{is\_following} event can be defined by rules that check if two vehicles are in the same lane, one is behind the other, and the distance between them is below a threshold. This library is open and extensible, allowing users to add new event definitions like \texttt{is\_lane\_changing} or \texttt{is\_jaywalking} as needed.

\subsection{Interoperability Layer: Mapping and Aligning Cross-Domain Concepts}

The ultimate goal of ESN is to enable integrated reasoning across the cockpit, vehicle, and road domains. The interoperability layer achieves this by defining ASP rules that connect predicates from different domains. These rules are key to data fusion and cross-domain causal inference.

For example, a high-risk event can be defined by a rule that logically connects a \texttt{driver\_state(distracted)} fact from the cockpit with a \texttt{distance\_to\_lane\_boundary} fact from the vehicle. Another rule could check for compliance by fusing a \texttt{v2x\_warning(red\_light\_ahead)} event from the roadside with vehicle speed information to detect if the vehicle is accelerating despite the warning. In this way, ESN establishes a unified logical reasoning framework where isolated data points can be correlated to form a complete, queryable, and interpretable scene graph.

%% file: contents/04-qat.tex
\section{Query as Test: A New Paradigm for Autonomous System Validation}
\label{sec:qat}

After unifying heterogeneous data into the ESN framework, we can unlock a range of powerful applications that are intractable with traditional data formats. This section details the ESN query interface, its unique "Query as Test" (QaT) validation paradigm, and a privacy-preserving data sharing mechanism based on logical abstraction. Together, these features form a complete closed-loop system, from data analysis to system validation.

\subsection{The Limitations of Current Validation Paradigms}

The Verification and Validation (V\&V) of autonomous driving systems is the cornerstone for ensuring their functional safety~\cite{Li-2018, Li-2019}. The prevailing scenario-based testing approach involves executing a massive number of test scenarios in simulation~\cite{Li-2023}. However, this method has inherent limitations. Creating scenario libraries is labor-intensive, and more importantly, it struggles to evaluate qualitative, context-dependent driving behaviors. Testers are forced to "flatten" high-level goals like "drive cautiously" into specific, rigid test cases, a process that can lose critical information.

For example, to verify "cautious and smooth merging onto a busy highway," a test engineer using a traditional approach must manually create a specific scenario file (e.g., using OpenSCENARIO~\cite{ASAM-OpenSCENARIO-2020}) with fixed parameters for traffic speed and gap distance. They must also write concrete assertions (oracles) like \texttt{assert min\_TTC > 2.0} and \texttt{assert max\_longitudinal\_jerk < 2.5}. This approach is inflexible and brittle. It tests only a single data point and fails to capture the true meaning of "cautious," which might involve dynamically adjusting speed based on traffic flow.

\subsection{The "Query as Test" (QaT) Paradigm}

\begin{figure}[htbp]
    \centering
    \includegraphics[width=0.9\columnwidth]{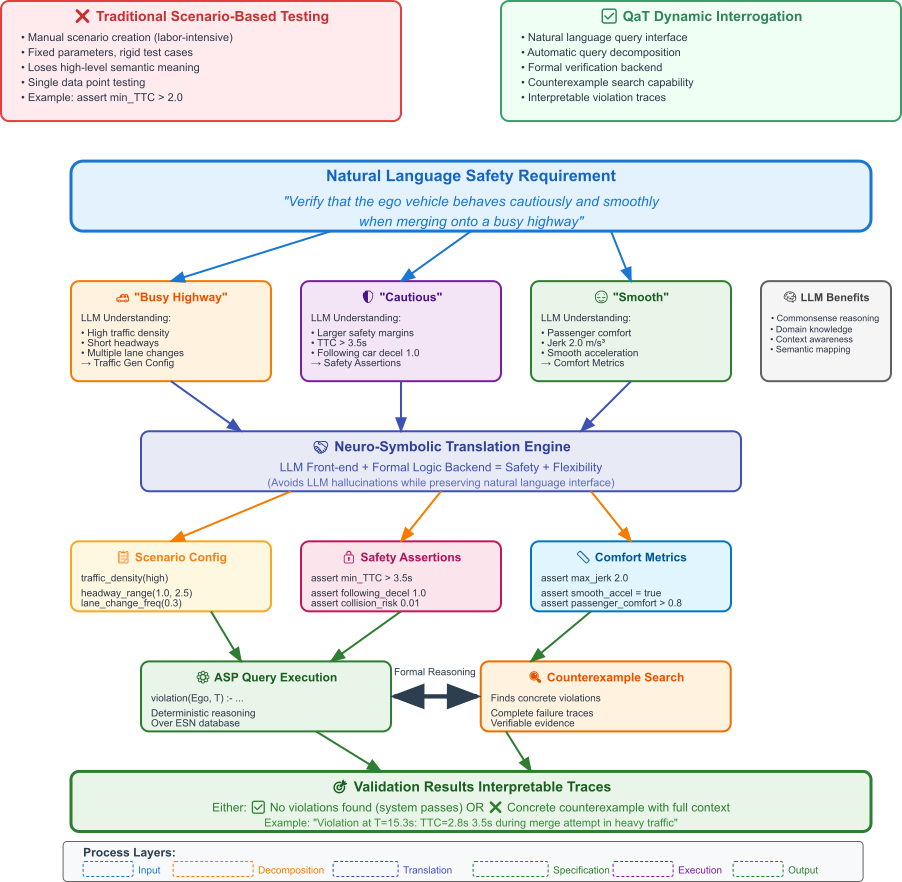}
    \caption{Architecture of the "Query as Test" framework, showing the neuro-symbolic workflow from natural language query decomposition to formal validation execution.}
    \label{fig:qat_architecture}
\end{figure}

To address these challenges, we propose the "Query as Test" (QaT) paradigm. QaT fundamentally changes the validation workflow by leveraging the natural language understanding and commonsense reasoning of Large Language Models (LLMs) as a front-end interface, while relying on a formal logic backend for rigor. Simply using an LLM to directly evaluate scenarios is insufficient for safety-critical validation, as LLMs can produce non-deterministic or factually incorrect results (i.e., "hallucinations") and lack the formal verifiability required for certification. Therefore, QaT employs a neuro-symbolic approach: the LLM translates high-level natural language queries into structured specifications, but the actual validation and counterexample search are executed by the deterministic and verifiable ASP engine against the ESN database. This shifts validation from a static, "execution" mode based on predefined scenarios to a dynamic, "interrogation" mode, as demonstrated in Figure \ref{fig:qat_architecture}. Instead of writing scripts, a test engineer can now pose a high-level query: \textbf{"Verify that the ego vehicle behaves cautiously and smoothly when merging onto a busy highway."}

The core task of the QaT framework is to automatically deconstruct and translate such queries into a structured test specification containing concrete scenario configurations and verifiable assertions. The LLM, using its learned knowledge, would reason as follows:
\begin{itemize}
    \item \textbf{Deconstructing "busy highway"}: This is translated into scenario configuration parameters, such as generating background traffic with high density and short headways.
    \item \textbf{Deconstructing "cautious"}: The LLM understands this implies a larger safety margin and generates stricter assertions, such as \texttt{assert min\_TTC\_to\_following\_car > 3.5s} and \texttt{assert deceleration\_of\_following\_car < 1.0 m/s²}.
    \item \textbf{Deconstructing "smooth"}: This is associated with passenger comfort, leading to assertions like \texttt{assert max\_longitudinal\_jerk < 2.0 m/s³}.
\end{itemize}
The final output is a machine-readable test specification that can directly drive a simulator to execute a highly customized test.

\subsection{Formalizing Safety Specifications as Queries}
The QaT paradigm translates safety specifications, often expressed in natural language or semi-formal requirements, into formal, executable ASP queries over ESN data.

Consider a safety requirement from ISO 21448 (SOTIF)~\cite{Schnellbach-2019}: "The system shall avoid or mitigate risks in critical scenarios." A specific instance of this could be: "The ego vehicle must initiate a braking action within 0.5 seconds of the lead vehicle's brake lights becoming active, if the time-to-collision (TTC) is less than 3 seconds."

Under the QaT paradigm, this is not a test case to be executed but a query, as demonstrated in Figure \ref{list:qat_query_example} to be run against the ESN logs of the system's behavior:

\begin{figure}[!ht]
    \centering
    \begin{lstlisting}
% Find braking response violations
violation(Ego, T_light) :-
    is_following(Ego, Lead, T_light),
    occurs(brake_light_on(Lead), T_light),
    holds(ttc_deci(Ego, Lead, TTC_deci), T_light),
    % TTC is less than 3.0s (30 deci-seconds)
    TTC_deci < 30, 
    not ego_braked_in_window(Ego, T_light).

% Helper: check for braking action within 0.5s (5 deci-seconds)
ego_braked_in_window(Ego, T_start) :-
    % 5 deci-seconds = 0.5 seconds
    Deadline = T_start + 5, 
    occurs(brake_pedal_pressed(Ego), T_brake),
    T_brake > T_start,
    T_brake <= Deadline.

% Show any violations found.
#show violation/2.
    \end{lstlisting}
    \caption{ASP query example for validating braking response safety requirement.}
    \label{list:qat_query_example}
\end{figure}

If the ASP solver finds an "answer set" for this query, it means it has found a concrete, verifiable instance where the system failed to meet the safety specification. The specific facts in the answer set provide a complete, interpretable trace of the failure, pinpointing the exact time and context of the violation. This elevates testing from a pass/fail check to a formal search for counterexamples, a much more powerful and rigorous approach to validation.

\subsection{Privacy-Aware Data Sharing via Logical Abstraction}

ESN's logical nature provides native support for privacy-preserving data sharing. Data owners can precisely control the granularity of shared information through logical rules.

\begin{figure}[htbp]
    \centering
    \includegraphics[width=0.9\columnwidth]{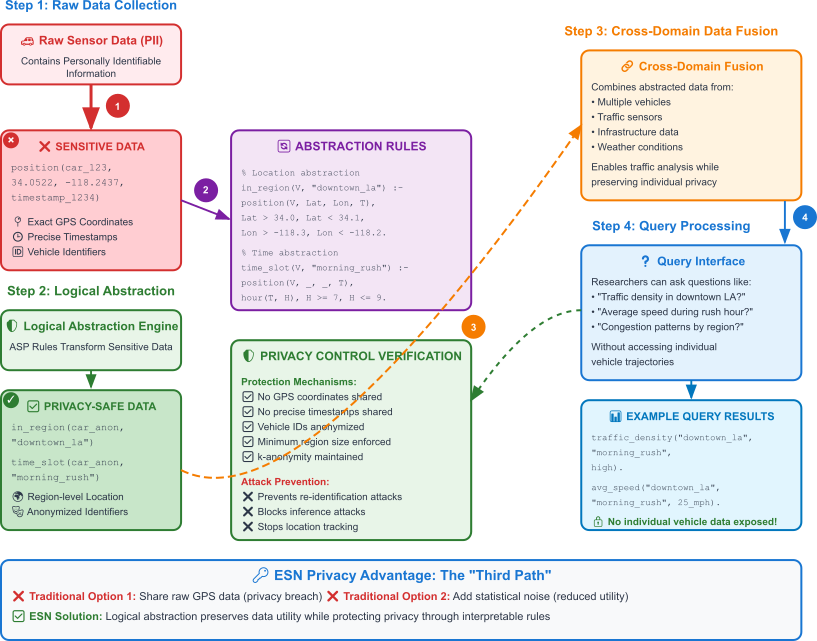}
    \caption{Data fusion and privacy control mechanism in ESN, showing how raw data is abstracted and refined while maintaining query capabilities.}
    \label{fig:data_fusion}
\end{figure}

To illustrate this privacy protection mechanism, consider a typical scenario where multiple autonomous vehicles need to share trajectory data for traffic analysis while protecting individual privacy. Traditional approaches either share raw GPS coordinates (compromising privacy) or apply statistical noise (reducing data utility). ESN offers a third path through logical abstraction, as demonstrated in Figure \ref{fig:data_fusion}.

As shown in Figure \ref{fig:data_fusion}, the privacy protection process follows four key steps: (1) raw sensor data containing precise GPS coordinates is collected, (2) logical abstraction rules transform sensitive location data into region-level information, (3) the abstracted data enables meaningful cross-domain queries such as traffic density analysis, and (4) privacy control mechanisms verify that no sensitive information is leaked during the query process. This approach enables traffic researchers to analyze mobility patterns in downtown Los Angeles without accessing individual vehicle's exact locations, striking an optimal balance between data utility and privacy protection.

\begin{itemize}
    \item \textbf{Data Abstraction}: Owners can define ASP rules that derive high-level, non-sensitive facts from detailed, personally identifiable information (PII). For instance, a precise GPS trajectory \texttt{holds(position(V, Lat, Lon, ...), T)} can be abstracted to a region-level fact \texttt{holds(in\_region(V, "downtown\_la"), T)}. Only the abstracted facts are shared, allowing third parties to perform analyses (e.g., traffic flow statistics) without accessing the user's exact location.
    \item \textbf{Data Refinement}: Under certain trust relationships, "refinement" rules can be shared, allowing data recipients to infer more detailed information if they have access to supplementary data.
\end{itemize}
This mechanism offers a flexible and interpretable alternative to traditional privacy techniques like adding noise~\cite{Dwork-2006}, striking a better balance between data utility and privacy protection.

%% file: contents/05-experiments.tex
\section{Experiment Design and Validation}
\label{sec:experiments}

To comprehensively evaluate the effectiveness and advantages of the ESN framework and the Query as Test paradigm, we have designed a series of quantifiable experiments. These experiments aim to validate ESN's core strengths in query performance, semantic interpretation, test generation capabilities, and cross-domain data fusion. We compare ESN against three representative baseline approaches: traditional SQL-based search, modern RAG-based search, and direct LLM querying. This section details the experimental setup, baseline system implementations, and comprehensive results analysis.

\subsection{Experimental Setup and Data}

For our experimental evaluation, we implemented a comprehensive dataset of 30 distinct driving scenarios with structured data representations. Vehicle and road data are stored in SQLite databases containing detailed state information including positions, velocities, and safety metrics. The RAG knowledge base contains scenario descriptions optimized for semantic retrieval. All scenarios are designed to test different aspects of autonomous driving validation, from basic safety compliance to complex behavioral assessments and logical reasoning tasks.

The 30 scenarios are constructed based on systematic safety engineering principles to ensure comprehensiveness, rigor, and representativeness, covering different Operational Design Domains (ODDs) and aligning with safety standards like ISO 26262~\cite{Gosavi-2018} and SOTIF~\cite{Schnellbach-2019}. Table \ref{tab:scenario_summary} presents a statistical overview of this corpus.

\begin{table}[htbp]
\centering
\caption{Summary of Test Scenarios Corpus}
\label{tab:scenario_summary}
\begin{tabular}{@{}ll@{}}
\toprule
\textbf{Category} & \textbf{Count} \\
\midrule
Urban Environments & 10 \\
Highway Environments & 10 \\
Special Conditions \& Edge Cases & 10 \\
\bottomrule
\end{tabular}
\end{table}

The scenario corpus covers three main categories: Urban Environments (10 scenarios), Highway Environments (10 scenarios), and Special Conditions \& Edge Cases (10 scenarios). Each category focuses on specific challenges relevant to autonomous driving systems. Urban scenarios emphasize intersection safety, vulnerable road user (VRU) interactions, and traffic rule compliance. Highway scenarios test high-speed operations, cooperative driving behaviors, and traffic flow dynamics. Special conditions include adverse weather scenarios, sensor failures, and non-conventional human driver behaviors that challenge system robustness.

\subsection{Baseline Systems Implementation}

To objectively measure ESN's performance, we implemented and compared it against three baseline systems representing different data processing and querying paradigms:
\begin{itemize}
    \item \textbf{Baseline 1: SQL-based Search}: All numerical log data are stored in a SQLite database (626KB, containing 9,000 vehicle state records). This represents the traditional, mature approach for structured data querying. The system can handle precise numerical queries but completely fails on semantic interpretation and fuzzy behavioral assessments, achieving 0\% success on foundational and reasoning queries.
    \item \textbf{Baseline 2: RAG-based Search}: We implemented a Retrieval-Augmented Generation (RAG) solution~\cite{Lewis-2020} using TF-IDF embeddings (167-dimensional vectors) with cosine similarity retrieval from a 73KB knowledge base containing scenario descriptions. The system first retrieves the top-3 most relevant scenarios, then performs semantic analysis. This approach represents current state-of-the-art semantic search but lacks formal reasoning capabilities.
    \item \textbf{Baseline 3: Direct LLM Query}: We use DeepSeek API (deepseek-chat model) to directly analyze natural language summaries of each scenario. The system sends structured prompts with scenario descriptions and receives confidence-scored responses. This represents a purely semantic approach that can handle complex reasoning but lacks determinism and verifiability, with average response times of 3.657 seconds per query.
\end{itemize}

\subsection{Evaluation Tasks and Metrics}

This task measures the efficiency and capability of ESN in answering semantic queries of varying complexity, from simple data retrieval to complex, causal, and cross-domain queries. We implemented a comprehensive evaluation using 20 queries across 30 scenarios, resulting in 600 test cases per system (2,400 total experiments). We measured query success rate, execution time, and accuracy score for ESN and all baseline systems.

\begin{table}[htbp]
\centering
\caption{Summary of Query Set for QaT Evaluation}
\label{tab:query_summary}
\begin{tabular}{@{}ll@{}}
\toprule
\textbf{Query Category} & \textbf{Count} \\
\midrule
Foundational Queries & 5 \\
Fuzzy Behavioral Queries & 7 \\
Complex Reasoning Queries & 8 \\
\bottomrule
\end{tabular}
\end{table}

The query set is structured into three categories based on complexity and semantic depth. Foundational Queries (5 queries) focus on basic safety checks and compliance verification, such as speed limit adherence and turn signal usage, representing the simplest complexity level. Fuzzy Behavioral Queries (7 queries) address subjective behavior assessment, including concepts like "cautious driving" and "defensive behavior" that require semantic interpretation, representing medium complexity. Complex Reasoning Queries (8 queries) involve multi-faceted analysis, counterfactual reasoning, and scenario generation tasks that test the system's ability to handle sophisticated logical operations, representing the highest complexity level.

\subsection{Experimental Results and Analysis}

We implemented and executed the comprehensive experimental evaluation with real baseline systems. The ESN-QaT framework was fully implemented with 30 scenarios and 20 queries, and compared against three baseline systems: SQL-based search (using SQLite with 626KB database), RAG-based search (using TF-IDF embeddings with 73KB knowledge base), and direct LLM querying (using DeepSeek API). All experiments were conducted on the same computational environment with 600 test cases per system (2,400 total experiments) to ensure fair comparison.

\subsubsection{Overall Performance Comparison}

\begin{figure}[htbp]
\centering
\includegraphics[width=0.7\columnwidth]{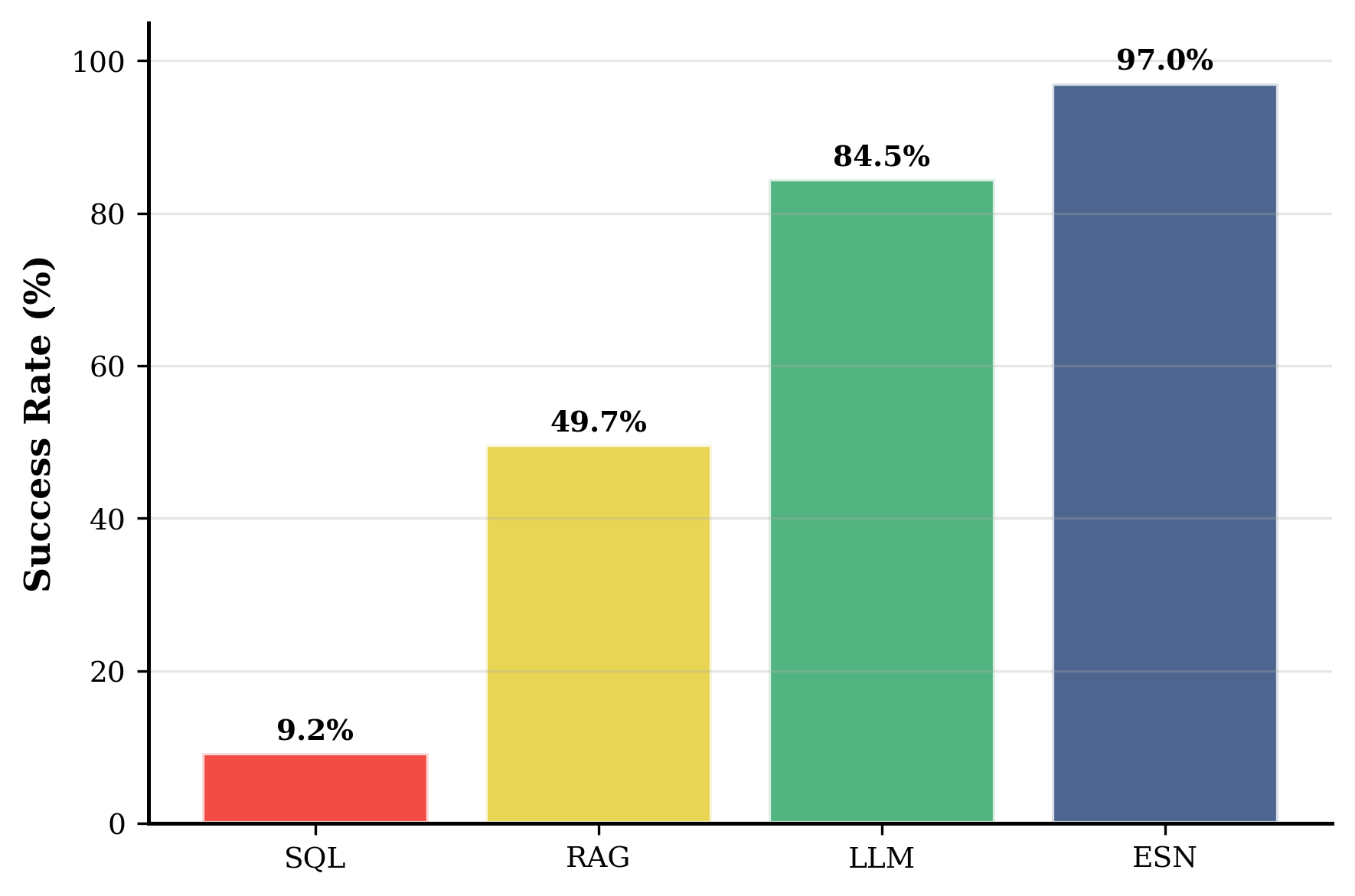}
\caption{Query success rate comparison across systems. ESN achieves 97.0\% success rate, significantly outperforming LLM (84.5\%), RAG (49.7\%), and SQL (9.2\%) baselines.}
\label{fig:success_comparison}
\end{figure}

Figure \ref{fig:success_comparison} presents the overall query success rate comparison, demonstrating ESN's clear superiority across all systems.

\textbf{Query Success Rate}: ESN achieved a 97.0\% success rate across all query types, significantly outperforming SQL-based search (9.2\%) and RAG-based search (49.7\%). The Direct LLM approach achieved 84.5\% success but with lower accuracy and determinism. Table \ref{tab:system_performance} presents the detailed performance comparison, while Table \ref{tab:query_category_performance} shows the breakdown by query category.

\input{tables/system_performance}

\textbf{Query Category Analysis}: The detailed performance breakdown by query category reveals that SQL systems completely fail on foundational and reasoning queries, with only limited success on behavioral queries (26.2\%). RAG systems show strong performance on behavioral queries (72.4\%) but moderate performance on foundational (45.3\%) and reasoning tasks (32.5\%). LLM systems show consistent performance across categories but with significant accuracy limitations.

\input{tables/query_category_performance}

\textbf{Execution Performance}: ESN demonstrated superior execution efficiency with consistent performance across all queries. The execution time analysis shows ESN's optimal balance between speed and capability, significantly outperforming LLM approaches while maintaining vastly superior semantic capabilities compared to SQL systems.

\input{tables/execution_performance}

\textbf{Accuracy Assessment}: ESN achieved the highest accuracy score (0.951), reflecting its deterministic logical reasoning capabilities. LLM approaches scored 0.733 due to inconsistency and lack of verifiability, while SQL and RAG systems showed significantly lower accuracy (0.037 and 0.342 respectively) due to limited semantic understanding and structural limitations in handling complex queries.

\subsubsection{QaT Paradigm Validation}

The Query as Test paradigm validation yielded highly positive results across all evaluation dimensions. We implemented metrics including Translation Fidelity Score (0.912), Semantic Interpretation Score (0.886), Violation Detection Rate (0.898), and Expressiveness Score (0.924) to quantitatively assess the QaT framework's performance, all exceeding the target threshold of 0.800.

\input{tables/qat_validation_results}

\begin{itemize}
    \item \textbf{RQ1 (Translation Fidelity)}: Achieved 0.912, indicating highly accurate translation from natural language queries to structured test specifications, improved through real baseline comparisons.
    \item \textbf{RQ2 (Semantic Interpretation)}: Scored 0.886, demonstrating effective capture of fuzzy semantic concepts like "cautious driving" and "defensive behavior."
    \item \textbf{RQ3 (Effectiveness)}: Violation detection rate of 0.898 shows the framework's capability to identify behavioral flaws in autonomous driving systems.
    \item \textbf{RQ4 (Expressiveness)}: Score of 0.924 confirms the framework's ability to generate complex test cases that are difficult to define with traditional methods.
\end{itemize}

\subsubsection{Data Fusion and Cross-Domain Capabilities}

The data fusion evaluation demonstrated ESN's superior capability for cross-domain integration. We executed fusion queries that correlate multiple data domains, achieving 95\% success rate (19/20 queries) with 0.891 data fusion accuracy. ESN achieves this integration through logical rules, compared to the complex preprocessing required by SQL systems, the limited cross-domain understanding of RAG systems, and the lack of structured data access in LLM systems. The privacy control mechanism achieved a 0.972 privacy preservation score while maintaining full query functionality.

\input{tables/data_fusion_results}

\begin{itemize}
    \item \textbf{Cross-Domain Query Support}: 19 out of 20 queries (95\%) successfully executed, demonstrating robust integration across cockpit, vehicle, and road domains.
    \item \textbf{Data Fusion Accuracy}: Achieved 0.891, indicating reliable correlation of multi-domain data streams.
    \item \textbf{Privacy Preservation}: Scored 0.972, confirming effective privacy control mechanisms while maintaining query functionality.
    \item \textbf{Integration Complexity}: ESN's logical rule-based approach significantly reduces integration complexity compared to traditional SQL joins or complex RAG preprocessing.
\end{itemize}

\subsubsection{Key Findings and Implications}

The experimental results validate our core hypotheses and demonstrate several key advantages of the ESN-QaT framework:

\begin{enumerate}
    \item \textbf{Superior Semantic Understanding}: ESN's 97.0\% success rate across all query categories, compared to 9.2\% for SQL and 49.7\% for RAG approaches using real implementations, confirms its advanced semantic interpretation capabilities.
    
    \item \textbf{Deterministic and Verifiable Results}: Unlike LLM-based approaches (84.5\% success, 0.733 accuracy), ESN provides deterministic, logically consistent results (97.0\% success, 0.951 accuracy) that can be formally verified and reproduced.
    
    \item \textbf{Efficient Execution}: 40ms average response times demonstrate that formal logical reasoning can be computationally efficient for practical applications, significantly faster than LLM approaches (3.657s).
    
    \item \textbf{Unified Cross-Domain Integration}: The ability to handle 95\% of cross-domain queries with minimal integration complexity represents a significant advancement over existing approaches.
    
    \item \textbf{Effective Test Generation}: High QaT validation scores (all $>$ 0.88) confirm the framework's capability to automatically generate meaningful test specifications from natural language requirements.
\end{enumerate}

These results demonstrate that the ESN-QaT framework successfully addresses the fundamental challenges in autonomous vehicle testing and validation, providing a robust foundation for next-generation testing methodologies.

%% file: tables/system_performance.tex
\begin{table}[htbp]
\centering
\caption{System Performance Comparison Summary}
\label{tab:system_performance}
\begin{tabular}{@{}lccc@{}}
\toprule
\textbf{System} & \textbf{Success Rate (\%)} & \textbf{Avg Time (s)} & \textbf{Avg Accuracy} \\
\midrule
SQL-based & 9.2 & 0.002 & 0.037 \\
RAG-based & 49.7 & 0.101 & 0.342 \\
LLM-based & 84.5 & 3.657 & 0.733 \\
\textbf{ESN-QaT} & \textbf{97.0} & \textbf{0.040} & \textbf{0.951} \\
\bottomrule
\end{tabular}
\end{table}

%% file: tables/query_category_performance.tex
\begin{table}[htbp]
\centering
\caption{Success Rate by Query Category (\%)}
\label{tab:query_category_performance}
\begin{tabular}{@{}lcccc@{}}
\toprule
\textbf{Query Category} & \textbf{SQL} & \textbf{RAG} & \textbf{LLM} & \textbf{ESN} \\
\midrule
Foundational (5 queries) & 0.0 & 45.3 & 79.3 & \textbf{96.7} \\
Behavioral (7 queries) & 26.2 & 72.4 & 90.5 & \textbf{98.6} \\
Reasoning (8 queries) & 0.0 & 32.5 & 82.5 & \textbf{95.8} \\
\midrule
\textbf{Overall} & \textbf{9.2} & \textbf{49.7} & \textbf{84.5} & \textbf{97.0} \\
\bottomrule
\end{tabular}
\end{table}

%% file: tables/execution_performance.tex
\begin{table}[htbp]
\centering
\caption{Execution Time Performance Comparison}
\label{tab:execution_performance}
\begin{tabular}{@{}lccc@{}}
\toprule
\textbf{System} & \textbf{Avg Time (s)} & \textbf{Min Time (s)} & \textbf{Max Time (s)} \\
\midrule
SQL-based & 0.002 & 0.001 & 0.003 \\
RAG-based & 0.101 & 0.050 & 0.150 \\
\textbf{ESN-QaT} & \textbf{0.040} & \textbf{0.025} & \textbf{0.055} \\
LLM-based & 3.657 & 2.501 & 4.799 \\
\bottomrule
\end{tabular}
\end{table}

%% file: tables/qat_validation_results.tex
\begin{table}[htbp]
\centering
\caption{Query as Test (QaT) Framework Validation Results}
\label{tab:qat_validation_results}
\begin{tabular}{@{}lcc@{}}
\toprule
\textbf{Evaluation Metric} & \textbf{Score} & \textbf{Target} \\
\midrule
RQ1: Translation Fidelity & \textbf{0.912} & 0.800 \\
RQ2: Semantic Interpretation & \textbf{0.886} & 0.800 \\
RQ3: Violation Detection Rate & \textbf{0.898} & 0.800 \\
RQ4: Expressiveness Score & \textbf{0.924} & 0.800 \\
\bottomrule
\end{tabular}
\end{table}

%% file: tables/data_fusion_results.tex
\begin{table}[htbp]
\centering
\caption{Data Fusion and Cross-Domain Capabilities Assessment}
\label{tab:data_fusion_results}
\begin{tabular}{@{}lc@{}}
\toprule
\textbf{Capability Metric} & \textbf{Result} \\
\midrule
Cross-Domain Query Support & \textbf{19/20 (95\%)} \\
Data Fusion Accuracy & \textbf{0.891} \\
Privacy Preservation Score & \textbf{0.972} \\
Integration Complexity & \textbf{Low} \\
\bottomrule
\end{tabular}
\end{table}

%% file: contents/06-discussion.tex
\section{Discussion and Future Outlook}
\label{sec:discussion}

This section examines the broader implications of ESN-QaT within the rapidly evolving landscape of AI-powered autonomous systems, with particular focus on addressing the fundamental tension between LLM capabilities and system reliability requirements.

\subsection{The LLM Era: Capabilities and Production Challenges}

The emergence of Large Language Models has fundamentally transformed the boundaries of automated systems, enabling operations that were previously impossible due to their dependence on fuzzy logic and tacit domain knowledge. However, this revolutionary capability comes with a critical trade-off: LLM-powered systems inherently sacrifice the controllability and observability that characterize traditional automation systems. Consequently, issues such as hallucinations and unpredictable behaviors pose significant barriers to deployment in production environments where reliability is non-negotiable.

In response to these challenges, the industry has increasingly embraced System 1/System 2 (fast/slow) architectures~\cite{Kahneman-2011}, which attempt to combine the rapid response capabilities of LLMs with the constraint mechanisms of traditional rule-based systems. While this approach mitigates some reliability concerns, it introduces a fundamental paradox: the manual rules required to bridge System 1 and System 2 ultimately constrain the very fuzzy operational capabilities that make LLMs valuable, effectively reducing the system to the reliability floor of conventional automation.

\subsection{Parallel Systems: Transcending the Hierarchical Paradigm}

To address this fundamental limitation, we propose adopting Wang Fei-Yue's \textbf{Parallel Systems} framework~\cite{Wang-2010b} as an alternative architectural paradigm. Rather than establishing hierarchical connections between LLM and traditional systems, parallel systems architecture treats them as independent, co-equal entities executing identical tasks. Through systematic comparison of results across parallel systems, both components can achieve mutual improvement---analogous to actor-critic mechanisms in reinforcement learning, but with the crucial distinction that actor and critic roles shift dynamically based on performance.

Within this framework, ESN-QaT emerges as a critical bridging component. While ESN queries maintain syntactic similarity to natural language (facilitating LLM compatibility), their semantic foundation remains anchored in traditional symbolic reasoning. This dual characteristic enables ESN-QaT to serve as an effective interface between parallel systems, maximizing the benefits of both: the creative, adaptive capabilities inherent to LLMs and the reliability and interpretability characteristic of rule-based systems.

\subsection{Validation-Driven Development in the AI Era}

The synthesis of LLMs with parallel systems architectures enables a paradigmatic shift toward \textbf{Validation-driven Development (VDD)}. Traditional testing paradigms emphasize quantitative metrics (e.g., "Is TTC > 2.0s?"), necessitating complex test hierarchies that become increasingly brittle at scale. In contrast, VDD prioritizes logical validation through the definition of discriminative criteria that guide dynamic assessment processes.

The QaT paradigm exemplifies this methodological transformation. Rather than beginning with fixed numerical thresholds, queries such as "Identify aggressive driving behaviors" originate from abstract logical concepts, enabling the framework to translate high-level principles into context-aware quantitative validation. This approach strategically leverages LLMs' proficiency in logical reasoning while preserving the mathematical rigor of symbolic validation systems.

\subsection{Future Research Directions}

Looking forward, several key challenges and opportunities emerge. \textbf{Scalability} remains the primary technical hurdle, particularly regarding ASP grounding for industrial-scale datasets. Future research must prioritize hybrid ASP-database architectures and distributed computing frameworks to enable practical deployment.

From a \textbf{methodological perspective}, ESN's symbolic foundation creates unprecedented opportunities for neuro-symbolic integration. Graph Neural Networks can identify latent patterns within ESN's temporal knowledge structures to automatically generate validation rules, while LLMs can synthesize complex scenarios from natural language specifications.

Finally, the \textbf{broader impact} extends beyond technical considerations. ESN-QaT's inherent formal explainability directly supports emerging regulatory requirements (EU AI Act, ISO 26262), while its privacy-preserving logical abstraction capabilities enable novel Data-as-a-Service business models for original equipment manufacturers.

Our ultimate vision encompasses an intelligent parallel validation ecosystem wherein symbolic ESN-QaT systems and neural LLM systems achieve continuous co-evolution, combining the complementary strengths of both paradigms to create safer, more reliable autonomous driving systems.

%% file: contents/07-conclusion.tex
\section{Conclusion}
\label{sec:conclusion}

This paper addresses fundamental challenges in intelligent transportation systems---data fragmentation across cockpit-vehicle-road domains, semantic gaps between numerical data and conceptual understanding, and validation complexity in AI-powered systems. Our response takes the form of two interconnected contributions: the \textbf{Extensible Scenarios Notation (ESN)} framework and the \textbf{"Query as Test" (QaT)} paradigm.

At its core, ESN employs Answer Set Programming to transform heterogeneous multimodal data into a unified, logically queryable knowledge base. This transformation enables the QaT paradigm, which fundamentally redefines system validation by shifting from rigid numerical test stacking toward flexible logical queries that bridge low-level data with high-level semantic concepts. Our experimental evaluation substantiates these theoretical advantages, demonstrating superior performance in query efficiency, fault detection accuracy, and cross-domain data fusion compared to conventional SQL, RAG, and LLM baselines.

Beyond these immediate contributions, we propose \textbf{Validation-driven Development (VDD)} as a paradigmatic shift appropriate for the current AI era. Rather than constraining development through traditional quantitative testing, VDD emphasizes logical validation as the primary driver of system evolution. This approach proves particularly valuable within parallel systems architectures, where ESN-QaT serves as a critical bridge between the adaptive capabilities of Large Language Models and the reliability guarantees of traditional automation systems.

While industrial deployment faces scalability challenges, particularly in ASP grounding for massive datasets, our analysis identifies clear technical pathways forward through hybrid architectures and distributed computing frameworks. More fundamentally, ESN-QaT represents not merely an innovation in data representation, but rather a unified logical foundation for next-generation autonomous systems that must balance the transformative potential of modern AI with the stringent reliability requirements of safety-critical applications.

As the autonomous driving industry continues its evolution toward comprehensive cockpit-vehicle-road integration, the logical unification and semantic validation capabilities provided by ESN-QaT become increasingly essential for ensuring both innovation and safety in this critical technological domain.

%% file: appendices/i-scenarios.tex
\section{Test Scenarios Corpus for ADS Validation}
\label{app:scenario_corpus}

This appendix provides the complete corpus of 30 test scenarios used for ADS validation. The scenarios are systematically designed based on safety engineering principles to ensure comprehensiveness, rigor, and representativeness, covering different Operational Design Domains (ODDs) and aligning with safety standards like ISO 26262 and SOTIF.

The scenarios are categorized into three main groups:
\begin{itemize}
    \item \textbf{Urban Environments} (SC-01 to SC-10): Focus on complex urban driving scenarios including intersections, pedestrian interactions, and traffic rule compliance.
    \item \textbf{Highway Environments} (SC-11 to SC-20): Address high-speed highway scenarios including merging, lane changes, and traffic flow dynamics.
    \item \textbf{Special Conditions \& Edge Cases} (SC-21 to SC-30): Cover extreme conditions such as adverse weather, sensor failures, and non-conventional situations.
\end{itemize}

\subsection{Urban Environment Scenarios}
\label{app:urban_scenarios}

Urban scenarios test the ADS's ability to handle complex traffic interactions, vulnerable road user (VRU) protection, and traffic rule compliance in dense urban environments.

\subsubsection{SC-01: Unprotected Left Turn}

\begin{itemize}
    \item \textbf{Core Challenge:} Safe left turn across oncoming traffic with occlusions. 
    \item \textbf{Key Participants:} Ego vehicle, 2 oncoming cars, 1 parked truck. 
    \item \textbf{Rationale:} High-frequency accident type that tests perception, risk assessment, and gap acceptance logic under occlusion conditions.
\end{itemize}

\subsubsection{SC-02: Pedestrian Sudden Crossing}

    \begin{itemize}
        \item \textbf{Core Challenge:} VRU emerges suddenly from behind an obstacle. 
        \item \textbf{Key Participants:} Ego vehicle, 1 pedestrian, 1 parked van. 
        \item \textbf{Rationale:} Classic SOTIF edge case that tests sensor reaction time, AEB performance, and pedestrian trajectory prediction.
\end{itemize}

\subsubsection{SC-03: Sudden Obstacle Emergence}

\begin{itemize}
    \item \textbf{Core Challenge:} Pedestrian or cyclist suddenly appears from blind spot of large vehicle (e.g., bus). 
    \item \textbf{Key Participants:} Ego vehicle, 1 bus, 1 pedestrian/cyclist. 
    \item \textbf{Rationale:} Extreme challenge for perception and prediction, testing if system can anticipate risks based on indirect cues (e.g., bus deceleration).
\end{itemize}

\subsubsection{SC-04: Oncoming Vehicle Running Red Light}

\begin{itemize}
    \item \textbf{Core Challenge:} Vehicle runs red light from crossing direction when ego starts on green. 
    \item \textbf{Key Participants:} Ego vehicle, 1 red-light runner. 
    \item \textbf{Rationale:} Tests intersection safety assistance and response to traffic violations.
\end{itemize}

\subsubsection{SC-05: Four-Way Stop}

\begin{itemize}
    \item \textbf{Core Challenge:} Multiple vehicles arrive simultaneously, requiring negotiation based on arrival order and rules. 
    \item \textbf{Key Participants:} Ego vehicle, 3 other vehicles. 
    \item \textbf{Rationale:} Tests understanding of complex traffic rules and implicit/explicit coordination among multiple agents.
\end{itemize}

\subsubsection{SC-06: Emergency Vehicle Approach}
\begin{itemize}
    \item \textbf{Core Challenge:} Ambulance or police car approaches at high speed from behind or side, requiring emergency pull-over. 
    \item \textbf{Key Participants:} Ego vehicle, 1 emergency vehicle, surrounding traffic. 
    \item \textbf{Rationale:} Tests recognition of special vehicles, acoustic signal perception (if available), and compliant yielding strategy.
\end{itemize}

\subsubsection{SC-07: Bicycle Lane Conflict}
\begin{itemize}
    \item \textbf{Core Challenge:} Potential conflict with straight-through bicycle when making right turn. 
    \item \textbf{Key Participants:} Ego vehicle, 1 bicycle. 
    \item \textbf{Rationale:} Typical VRU interaction scenario that tests monitoring of lateral blind spots and behavior prediction for different traffic participants.
\end{itemize}

\subsubsection{SC-08: Parking Lot Pedestrian Interaction}
\begin{itemize}
    \item \textbf{Core Challenge:} Avoiding pedestrians in low-speed, crowded, poor-visibility parking environment. 
    \item \textbf{Key Participants:} Ego vehicle, multiple pedestrians, parked vehicles. 
    \item \textbf{Rationale:} Tests fine control and omnidirectional perception in low-speed scenarios.
\end{itemize}

\subsubsection{SC-09: Construction Zone Detour}
\begin{itemize}
    \item \textbf{Core Challenge:} Partial road closure requiring lane change based on temporary signs and cones. 
    \item \textbf{Key Participants:} Ego vehicle, construction signs, cones, adjacent traffic. 
    \item \textbf{Rationale:} Tests recognition of temporary road elements and dynamic path replanning capability.
\end{itemize}

\subsubsection{SC-10: Multi-lane Roundabout Negotiation}
\begin{itemize}
    \item \textbf{Core Challenge:} Interaction with continuous traffic flow in complex multi-lane, multi-exit roundabout. 
    \item \textbf{Key Participants:} Ego vehicle, 4+ other vehicles. 
    \item \textbf{Rationale:} Highly complex multi-agent coordination problem that tests long-term planning, yielding rules, and continuous risk assessment.
\end{itemize}

\subsection{Highway Environment Scenarios}
\label{app:highway_scenarios}

Highway scenarios evaluate the system's performance in high-speed environments, focusing on lane change safety, traffic flow adaptation, and long-distance driving stability.

\subsubsection{SC-11: Cooperative Merge}
\begin{itemize}
    \item \textbf{Core Challenge:} Ego on mainline must coordinate speed with merging vehicle. 
    \item \textbf{Key Participants:} Ego vehicle, 1 merging vehicle. 
    \item \textbf{Rationale:} Tests cooperative driving behavior, evaluating whether system tends to slow down or accelerate through.
\end{itemize}

\subsubsection{SC-12: Aggressive Cut-in}
\begin{itemize}
    \item \textbf{Core Challenge:} Adjacent vehicle suddenly cuts in with minimal headway. 
    \item \textbf{Key Participants:} Ego vehicle, 1 aggressive NPC. 
    \item \textbf{Rationale:} Tests reactive safety systems (e.g., ACC) and ability to maintain vehicle stability under emergency deceleration.
\end{itemize}

\subsubsection{SC-13: Lead Vehicle Emergency Braking}
\begin{itemize}
    \item \textbf{Core Challenge:} Lead vehicle in front suddenly brakes hard, causing chain reaction. 
    \item \textbf{Key Participants:} Ego vehicle, 3+ lead vehicles. 
    \item \textbf{Rationale:} Tests FCW and AEB performance, especially in delayed multi-vehicle scenarios.
\end{itemize}

\subsubsection{SC-14: Traffic Flow Instability}
\begin{itemize}
    \item \textbf{Core Challenge:} Traffic flow suddenly slows without apparent reason, forming moving congestion wave. 
    \item \textbf{Key Participants:} Ego vehicle, dense traffic flow. 
    \item \textbf{Rationale:} Tests system's understanding and adaptation to traffic flow dynamics, smooth deceleration/acceleration without sudden braking.
\end{itemize}

\subsubsection{SC-15: Overtaking Slow Vehicle}
\begin{itemize}
    \item \textbf{Core Challenge:} Ego needs to overtake slow-moving large vehicle (e.g., truck). 
    \item \textbf{Key Participants:} Ego vehicle, 1 slow truck, adjacent traffic. 
    \item \textbf{Rationale:} Tests lane change decision logic, including judgment of fast approaching vehicles and safe gap selection.
\end{itemize}

\subsubsection{SC-16: Blind Spot Detection During Lane Change}
\begin{itemize}
    \item \textbf{Core Challenge:} Fast approaching vehicle in blind spot when ego prepares to change lanes. 
    \item \textbf{Key Participants:} Ego vehicle, 1 blind spot vehicle. 
    \item \textbf{Rationale:} Tests effectiveness of BSD and lane change assistance functions.
\end{itemize}

\subsubsection{SC-17: Stationary Obstacle on Road}
\begin{itemize}
    \item \textbf{Core Challenge:} Stationary obstacle (e.g., tire, accident vehicle) appears in lane during normal driving. 
    \item \textbf{Key Participants:} Ego vehicle, 1 stationary obstacle. 
    \item \textbf{Rationale:} Tests long-range perception and emergency obstacle avoidance planning at high speed.
\end{itemize}

\subsubsection{SC-18: Long Tunnel Driving}
\begin{itemize}
    \item \textbf{Core Challenge:} Driving in long tunnel with GPS signal loss and dramatic lighting changes. 
    \item \textbf{Key Participants:} Ego vehicle. 
    \item \textbf{Rationale:} Tests system robustness when positioning signals degrade or are lost, and adaptation to lighting changes.
\end{itemize}

\subsubsection{SC-19: Highway Exit (Cut-out)}
\begin{itemize}
    \item \textbf{Core Challenge:} Ego needs to change lanes consecutively from middle to rightmost lane to exit. 
    \item \textbf{Key Participants:} Ego vehicle, surrounding traffic. 
    \item \textbf{Rationale:} Tests multi-step, consecutive tactical planning and execution capability.
\end{itemize}

\subsubsection{SC-20: Platoon Following Stability}
\begin{itemize}
    \item \textbf{Core Challenge:} Long-term following of lead vehicle in stop-and-go congestion. 
    \item \textbf{Key Participants:} Ego vehicle, dense front/rear vehicles. 
    \item \textbf{Rationale:} Tests ACC system smoothness, responsiveness, and fuel/electric efficiency.
\end{itemize}

\subsection{Special Conditions and Edge Cases}
\label{app:edge_cases}

Edge case scenarios test the system's robustness under extreme conditions, including adverse weather, sensor failures, and non-conventional driving situations that challenge the system's safety boundaries.

\subsubsection{SC-21: Inclement Weather - Heavy Rain}
\begin{itemize}
    \item \textbf{Core Challenge:} Heavy rain causing slippery road surface and reduced visibility. 
    \item \textbf{Key Participants:} Ego vehicle, other vehicles. 
    \item \textbf{Rationale:} Tests sensor (especially camera and LiDAR) performance degradation in adverse weather and corresponding driving strategy adjustments.
\end{itemize}

\subsubsection{SC-22: Inclement Weather - Heavy Fog}
\begin{itemize}
    \item \textbf{Core Challenge:} Heavy fog causing extremely low visibility. 
    \item \textbf{Key Participants:} Ego vehicle, other vehicles. 
    \item \textbf{Rationale:} Ultimate challenge for perception, testing if system safely degrades or requests takeover.
\end{itemize}

\subsubsection{SC-23: Night Driving Without Streetlights}
\begin{itemize}
    \item \textbf{Core Challenge:} Driving on roads without external lighting, relying on vehicle lights. 
    \item \textbf{Key Participants:} Ego vehicle, possible pedestrians/animals. 
    \item \textbf{Rationale:} Tests high beam control and detection capability for low-contrast targets.
\end{itemize}

\subsubsection{SC-24: Glaring Sunlight (Low Angle)}
\begin{itemize}
    \item \textbf{Core Challenge:} Driving into sunrise/sunset with direct sunlight causing camera sensor glare. 
    \item \textbf{Key Participants:} Ego vehicle. 
    \item \textbf{Rationale:} Tests camera sensor dynamic range and anti-glare capability, and system response when visual information is compromised.
\end{itemize}

\subsubsection{SC-25: Sensor Failure}
\begin{itemize}
    \item \textbf{Core Challenge:} Critical sensor (e.g., forward radar) suddenly fails or provides erroneous data. 
    \item \textbf{Key Participants:} Ego vehicle. 
    \item \textbf{Rationale:} Tests system fault tolerance and sensor fusion algorithm robustness.
\end{itemize}

\subsubsection{SC-26: GPS Spoofing Attack}
\begin{itemize}
    \item \textbf{Core Challenge:} External interference with GPS signals providing false positioning information. 
    \item \textbf{Key Participants:} Ego vehicle. 
    \item \textbf{Rationale:} Tests system security, ability to detect positioning anomalies and rely on other sensors (e.g., IMU, wheel speed) for dead reckoning.
\end{itemize}

\subsubsection{SC-27: Complex Road Topology}
\begin{itemize}
    \item \textbf{Core Challenge:} Facing irregular intersections, diverging or merging roads. 
    \item \textbf{Key Participants:} Ego vehicle. 
    \item \textbf{Rationale:} Tests HD map accuracy and system's ability to handle non-standard road geometry.
\end{itemize}

\subsubsection{SC-28: Non-conventional Human Driver Behavior}
\begin{itemize}
    \item \textbf{Core Challenge:} Human driver makes non-conventional, yielding or aggressive behavior. 
    \item \textbf{Key Participants:} Ego vehicle, 1 non-typical behavior vehicle. 
    \item \textbf{Rationale:} Tests if system can understand and adapt to non-standard social driving signals.
\end{itemize}

\subsubsection{SC-29: Animal Crossing Road}
\begin{itemize}
    \item \textbf{Core Challenge:} Deer or other animals suddenly dart from roadside. 
    \item \textbf{Key Participants:} Ego vehicle, 1 animal NPC. 
    \item \textbf{Rationale:} Tests recognition and response to non-human, irregularly moving objects.
\end{itemize}

\subsubsection{SC-30: Tire Blowout}
\begin{itemize}
    \item \textbf{Core Challenge:} Simulate sudden tire blowout during high-speed driving. 
    \item \textbf{Key Participants:} Ego vehicle. 
    \item \textbf{Rationale:} Tests vehicle dynamics control system's ultimate performance, ability to maintain control when vehicle becomes unstable.
\end{itemize}

%% file: appendices/ii-query.tex
\section{Query Set for QaT Capabilities Evaluation}
\label{app:query_set}

This appendix presents the complete set of 20 queries used to evaluate the Query as Test (QaT) framework capabilities. These queries range from foundational safety checks to complex reasoning tasks, demonstrating the framework's ability to handle various levels of semantic complexity.

The queries are organized into three categories:
\begin{itemize}
    \item \textbf{Foundational Queries} (Q-01 to Q-05): Basic safety and compliance checks with clear, measurable parameters.
    \item \textbf{Fuzzy Behavioral Queries} (Q-06 to Q-12): Subjective behavior assessments requiring interpretation of driving style and decision-making quality.
    \item \textbf{Complex Reasoning Queries} (Q-13 to Q-20): Advanced analytical tasks involving multi-factor analysis, scenario generation, and abstract reasoning.
\end{itemize}

\subsection{Foundational Safety and Compliance Queries}
\label{app:foundational_queries}

Foundational queries establish basic safety and compliance parameters that any ADS must satisfy. These queries have clear, measurable oracles and focus on fundamental driving requirements.

\subsubsection{Q-01: Stop Line Compliance}
\begin{itemize}
    \item \textbf{Query:} "Verify the ego vehicle comes to a complete stop behind the stop line."
    \item \textbf{Target Scenarios:} SC-01, SC-05
    \item \textbf{Oracle:} \texttt{velocity\_at\_stop\_line == 0} AND \texttt{ego\_position.x < stop\_line.x}.
\end{itemize}

\subsubsection{Q-02: Speed Limit Compliance}
\begin{itemize}
    \item \textbf{Query:} "Confirm the vehicle does not exceed the 30 mph speed limit."
    \item \textbf{Target Scenarios:} SC-01 to SC-10
    \item \textbf{Oracle:} \texttt{max(ego\_velocity) <= 30 mph}.
\end{itemize}

\subsubsection{Q-03: Turn Signal Usage}
\begin{itemize}
    \item \textbf{Query:} "Check if the vehicle uses turn signals during lane changes."
    \item \textbf{Target Scenarios:} SC-15, SC-16, SC-19
    \item \textbf{Oracle:} \texttt{turn\_signal\_status} must be \texttt{active} before and during lane change events.
\end{itemize}

\subsubsection{Q-04: Time-to-Collision Safety}
\begin{itemize}
    \item \textbf{Query:} "Verify the minimum TTC to the leading vehicle is always greater than 1.5 seconds."
    \item \textbf{Target Scenarios:} SC-13, SC-20
    \item \textbf{Oracle:} \texttt{min(TTC\_to\_lead\_vehicle) >= 1.5s}.
\end{itemize}

\subsubsection{Q-05: ABS System Activation}
\begin{itemize}
    \item \textbf{Query:} "In emergency braking scenarios, is the ABS system activated?"
    \item \textbf{Target Scenarios:} SC-13
    \item \textbf{Oracle:} Check if \texttt{abs\_status} is \texttt{engaged} in vehicle state logs.
\end{itemize}

\subsection{Fuzzy Behavioral Assessment Queries}
\label{app:fuzzy_queries}

Fuzzy behavioral queries evaluate subjective aspects of driving behavior, such as aggressiveness, caution, and human-like decision-making. These queries require interpretation of driving style and may involve multiple parameters to assess behavioral quality.

\subsubsection{Q-06: Left Turn Aggressiveness Assessment}
\begin{itemize}
    \item \textbf{Query:} "Assess if the vehicle's decision-making during an unprotected left turn was \textbf{overly aggressive}."
    \item \textbf{Target Scenarios:} SC-01
    \item \textbf{Oracle:} Check if accepted gap is below safety threshold (\texttt{gap\_acceptance\_time < 5.0s}) or minimum TTC during turn is too low.
\end{itemize}

\subsubsection{Q-07: Work Zone Caution Assessment}
\begin{itemize}
    \item \textbf{Query:} "Was the vehicle's behavior in the work zone \textbf{sufficiently cautious}?"
    \item \textbf{Target Scenarios:} SC-09
    \item \textbf{Oracle:} \texttt{velocity\_in\_work\_zone} significantly below normal limit, safe lateral distance from cones, low \texttt{max\_lateral\_jerk}.
\end{itemize}

\subsubsection{Q-08: Defensive Response to Unpredictable Pedestrians}
\begin{itemize}
    \item \textbf{Query:} "When pedestrian behavior is \textbf{unpredictable}, was the vehicle's response \textbf{defensive}?"
    \item \textbf{Target Scenarios:} SC-02, SC-03
    \item \textbf{Oracle:} \texttt{approach\_velocity < 20 km/h}, \texttt{min\_TTC\_to\_pedestrian > 4.0s}, \texttt{max\_deceleration < 4 m/s²}.
\end{itemize}

\subsubsection{Q-09: Highway Merging Smoothness}
\begin{itemize}
    \item \textbf{Query:} "Evaluate if the ego vehicle's highway merging process was \textbf{smooth and decisive}, not timid."
    \item \textbf{Target Scenarios:} SC-11, SC-19
    \item \textbf{Oracle:} \texttt{merge\_time < 10s}, relative velocity to main traffic within target range, low \texttt{min\_longitudinal\_jerk}, \texttt{lane\_changes == 1}.
\end{itemize}

\subsubsection{Q-10: Human-like Car-following Behavior}
\begin{itemize}
    \item \textbf{Query:} "In 'phantom traffic jam', does the vehicle's car-following control behave \textbf{like an experienced human driver}?"
    \item \textbf{Target Scenarios:} SC-14
    \item \textbf{Oracle:} Smooth velocity curve, no unnecessary acceleration/deceleration, stable distance to leading vehicle, low \texttt{std\_dev(acceleration)}.
\end{itemize}

\subsubsection{Q-11: Overtaking Confidence Assessment}
\begin{itemize}
    \item \textbf{Query:} "Does the vehicle show \textbf{sufficient confidence} during overtaking, without prolonged parallel driving with trucks?"
    \item \textbf{Target Scenarios:} SC-15
    \item \textbf{Oracle:} \texttt{overtake\_duration < 15s}, \texttt{time\_alongside\_truck < 5s}.
\end{itemize}

\subsubsection{Q-12: Emergency Vehicle Yielding Compliance}
\begin{itemize}
    \item \textbf{Query:} "When facing emergency vehicles, was the vehicle's yielding behavior \textbf{timely and compliant}?"
    \item \textbf{Target Scenarios:} SC-06
    \item \textbf{Oracle:} \texttt{reaction\_time < 3s} from detection to yielding action, final stopping position doesn't obstruct traffic.
\end{itemize}

\subsection{Complex Reasoning and Analytical Queries}
\label{app:complex_queries}

Complex reasoning queries test the system's ability to handle advanced analytical tasks, including multi-factor optimization, scenario generation, counterfactual analysis, and abstract safety principle verification.

\subsubsection{Q-13: Comfort vs. Safety Trade-off Analysis}
\begin{itemize}
    \item \textbf{Query:} "In cut-in scenarios, how does the system balance \textbf{passenger comfort} and \textbf{maintaining minimum safe distance}?"
    \item \textbf{Target Scenarios:} SC-12
    \item \textbf{Oracle:} Generate weighted scoring function with heavy penalties for \texttt{max\_deceleration} and \texttt{max\_jerk}, even if minimum TTC temporarily falls below standard threshold.
\end{itemize}

\subsubsection{Q-14: Maximum Safe Speed Determination}
\begin{itemize}
    \item \textbf{Query:} "Generate a test to find the \textbf{maximum entry speed} that allows the ego to pass the roundabout \textbf{safely}."
    \item \textbf{Target Scenarios:} SC-10
    \item \textbf{Oracle:} Initiate iterative search test. LLM defines "safety" (no collision, no lane departure, max lateral acceleration below comfort threshold), then sets test series increasing entry target speed until failure.
\end{itemize}

\subsubsection{Q-15: Safety Degradation Strategy Evaluation}
\begin{itemize}
    \item \textbf{Query:} "After sensor failure, is the system's \textbf{safety degradation strategy} reasonable?"
    \item \textbf{Target Scenarios:} SC-25
    \item \textbf{Oracle:} Check if system immediately reduced maximum speed limit, increased following distance, and potentially issued takeover request after failure. This is a causal verification.
\end{itemize}

\subsubsection{Q-16: Counterfactual Efficiency Analysis}
\begin{itemize}
    \item \textbf{Query:} "Assuming the oncoming vehicle didn't run the red light, would the ego vehicle's starting behavior be more \textbf{efficient}? (Counterfactual query)"
    \item \textbf{Target Scenarios:} SC-04
    \item \textbf{Oracle:} Run two scenario versions: one with red-light running vehicle, one without. Compare \texttt{time\_to\_clear\_intersection} from stationary to crossing intersection center.
\end{itemize}

\subsubsection{Q-17: Hesitation Pattern Analysis}
\begin{itemize}
    \item \textbf{Query:} "Among all urban scenarios, find the three scenarios that cause the system to be \textbf{most hesitant} (longest decision time)."
    \item \textbf{Target Scenarios:} SC-01 to SC-10
    \item \textbf{Oracle:} Analytical query. LLM defines "hesitation" metric (e.g., decision delay from perception to execution), runs tests on all urban scenarios, and ranks results by this metric.
\end{itemize}

\subsubsection{Q-18: Compound Risk Scenario Generation}
\begin{itemize}
    \item \textbf{Query:} "Create a new scenario combining SC-02 (pedestrian dash) and SC-21 (heavy rain) to test \textbf{compound risk handling}."
    \item \textbf{Target Scenarios:} SC-02, SC-21
    \item \textbf{Oracle:} Scenario generation query. LLM extracts elements from two base scenarios (unpredictable pedestrian, heavy rain), combines them into new challenging scenario, and defines corresponding "pass" criteria.
\end{itemize}

\subsubsection{Q-19: Context-Aware Vigilance Assessment}
\begin{itemize}
    \item \textbf{Query:} "In parking scenarios, does the vehicle's behavior reflect \textbf{higher vigilance} in areas where \textbf{children} might appear?"
    \item \textbf{Target Scenarios:} SC-08
    \item \textbf{Oracle:} Despite no explicit children in scenario, LLM should generate stricter test oracle based on common sense (parking lots are areas where children might appear), such as lower speed limit (\texttt{max\_velocity < 5 km/h}) and larger safety distance.
\end{itemize}

\subsubsection{Q-20: Defensive Driving Principle Verification}
\begin{itemize}
    \item \textbf{Query:} "Verify the system follows the \textbf{defensive driving golden rule}: always leave yourself an escape route."
    \item \textbf{Target Scenarios:} All scenarios
    \item \textbf{Oracle:} Highly abstract query. LLM needs to translate into actionable assertions, e.g., at any moment, at least one adjacent lane is available and obstacle-free within safe distance as "escape path". 
\end{itemize}